\documentclass[11pt]{article}

\usepackage[preprint]{acl}

\usepackage{times}
\usepackage{latexsym}
\usepackage{graphicx}
\usepackage{subcaption}
\usepackage{makecell}
\usepackage{array}
\usepackage{adjustbox}
\usepackage{enumitem}
\usepackage{mdframed}
\usepackage{amsmath}
\usepackage{amssymb}
\usepackage{pifont}
\usepackage{booktabs}
\usepackage{xspace}
\usepackage{multirow}
\usepackage[breakable]{tcolorbox}
\usepackage[textwidth=2cm, textsize=tiny]{todonotes}
\newcommand{\frameworkName}{\textsc{LeCat}\xspace}

\usepackage[T1]{fontenc}

\usepackage[utf8]{inputenc}

\usepackage{microtype}

\usepackage{inconsolata}

\usepackage{graphicx}

\usepackage{tabularx,booktabs,amsmath}

\title{\frameworkName: Measuring LLM Category Learning from Pretraining Data}

\usepackage{amssymb}  

\author{
  \textbf{Jimin Mun}\textsuperscript{$\heartsuit$} \quad
  \textbf{Maarten Sap}\textsuperscript{$\heartsuit$}\thanks{Equal advising contribution.} \quad
  \textbf{Yanai Elazar}\textsuperscript{$\clubsuit$}\footnotemark[1]
  \\
  \textsuperscript{$\heartsuit$}Carnegie Mellon University \quad
  \textsuperscript{$\clubsuit$}Bar-Ilan University
  \\
  \texttt{jmun@andrew.cmu.edu}
}

\begin{document}
\maketitle

\begin{abstract}
Humans know that \emph{airplanes are fast}, that this extends to \emph{red airplanes}, but that it does not hold for \emph{broken airplanes}. How do LLMs acquire category knowledge that humans learn and apply effortlessly?
We introduce \frameworkName, a two-part framework to analyze LLM's \underline{Le}arning of \underline{Cat}egory knowledge. First, we use \frameworkName to evaluate models on their ability to learn category knowledge, distinguishing between three cases: memorization of in-domain category knowledge, generalization to unseen instances, and ability to flexibly use knowledge in appropriate contexts. Then, we use \frameworkName to trace how this knowledge can be learned in pretraining from text that may encode category knowledge in various ways, such as verbatim repetition or exceptions, which we refer to as forms of ``knowledge expression''. We first assess open-source pretrained models using 199 curated category-feature pairs and analyze their training data. Our findings suggest that strengthening context and category-feature co-occurrence likelihood (PMI) improve learning for larger models and verbatim rule-like statements diminish flexible category learning.
To systematically isolate these effects, we then train a 68M model from scratch with precise control over the categorical information each sees in their training data, confirming that verbatim repetition reduces flexible learning but mixture of diverse knowledge expressions lead to more robust category learning.
Together, our findings highlight that while contextual flexibility is persistently difficult for models, \textit{how} category knowledge is expressed in pretraining greatly impacts models' acquisition of robust category knowledge.
\end{abstract}

\section{Introduction}
Category learning is fundamental to human intelligence \citep{murphy1994predictions,tenenbaum2006theory}, enabling generalization from sparse samples (e.g., \emph{airplanes are fast}) to large sets of unseen examples (e.g., \emph{red airplanes are fast}, \emph{blue airplanes are fast}). This capacity has been extensively studied in cognitive science and psychology \citep{banse2010development,markman2003category,rhodes2012cultural,wang2024maladaptive}, as well as in classical machine learning and computer vision \citep{jia2013visual,lake2015human}. Unlike factual knowledge, which is typically isolated (e.g., the Louvre opened on August 10, 1793), category membership licenses broader inference: the Louvre is a museum, so it likely shares characteristics with other museums \citep{bloom1964taxonomy,krathwohl2002revision,murphy2010category}. Yet this same inferential power can be harmful when applied inaccurately, as with social stereotypes \citep{beukeboom2019stereotypes,sherman1998stereotype}. 
Thus, for robust generalization, learners must also acquire exceptions and counter-stereotypes, and flexibly incorporate context.

In this work, we empirically investigate how language models learn to generalize and apply category knowledge under various contexts, focusing specifically on how pretraining data shapes model behavior.
While prior work has established that language models acquire factual or world knowledge from pretraining data \citep{eldan2023tinystories,hoffmann2022training} which has prompted extensive research on data curation \citep[e.g., ][]{longpre-etal-2024-pretrainers,parmar2024data,peng2025dataman} and knowledge injection \citep[e.g., ][]{chang2024large,lv2025inject}, how LMs acquire and use category knowledge from data is poorly understood. 
Clarifying how LLMs robustly acquire category knowledge would inform efforts to reduce harmful associations such as stereotypes and biases \citep{bai2025explicitly}, and promote more flexible, creative category representations that yield diverse outputs \citep{sassenberg2022flexibility,kenett2025role}.

We develop \frameworkName, a framework to analyze the \underline{Le}arning of \underline{Cat}egory knowledge and usage in LLM outputs and its training data. Focusing on LLM outputs, we develop log-probability measure, validated with LLM-as-a-judge, for category knowledge \emph{memorization} (i.e., generation of category knowledge in its basic form), \emph{generalization} (i.e., extension to invariant contexts), and \emph{contextual flexibility} (i.e., behavioral differences based on context).
For examining how training data informs LLM outputs, \frameworkName draws from cognitive science \citep{markman2003category} to measure the prevalence of five types of category knowledge expressions in data, including verbatim \textit{repetition} of category-feature associations (e.g., ``The airplanes are fast''), \emph{exceptions} (e.g., ``This airplane is not fast''), \emph{weakening} contexts (e.g., ``The airplanes with broken engines are not fast''), and more.

We first apply \frameworkName{} to open-data models of several sizes, Pythia~\citep{biderman2023pythia} and OLMo~3~\citep{olmo2025olmo}. At every scale, contextual flexibility remains difficult: models struggle to adjust which features they produce in context, especially when the context weakens or contradicts (i.e., is erroneous about) the category-feature association. In the pretraining data, the factors that support category learning shift with scale. Strengthening contexts and category-feature co-occurrence (measured by PMI) become more beneficial as models grow, with PMI correlating positively with learning only in larger models. Verbatim statements of category-feature associations, by contrast, hurt flexible knowledge use at every scale, possibly through memorization. Since pretraining data are heterogeneous, however, these findings remain correlational.

To test these trends causally and isolate individual expression types and paradigms such as curriculum ordering, we create a fully controlled pretraining corpus by adapting the TinyStories dataset \citep{eldan2023tinystories} by injecting two fictional category-feature relations, \emph{valins--poor} and \emph{pezzdens--fast}. We train a 68M parameter model from scratch on each expression type alone, a random mixture of all types, and a curriculum-ordered mixture. The results confirm the open-model finding that flexible category knowledge is difficult to learn but a diverse mixture improves while verbatim repetition reduces flexible category learning. 

Together, these findings show that pretraining data composition, not just exposure frequency, shapes category learning, that the effective composition shifts with model scale, and that contextual flexibility is persistently difficult for models. Our work thus points to practical implications for stereotypes and factual learning in LLMs. 

\section{Background \& Related Work}


Category learning is fundamental to human cognition, enabling classification, generalization, and inference \citep{tenenbaum2006theory,murphy2010category,barsalou2012human}. Humans acquire categories through diverse mechanisms—from explicit rules to implicit similarity detection \citep{minda2024single,murphy2004typicality}—and readily accommodate exceptions to typical category structure \citep{heffernan2021learning}. Crucially, human category representations are context-dependent: many categorical concepts are multi-dimensional and distributed across brain regions, allowing certain features to be more activated depending on context \citep{yee2012colorless,yee2016putting}. This flexibility supports accurate generalization, reduces stereotyping \citep{sassenberg2022flexibility}, and facilitates creative reasoning \citep{kenett2025role}.

Prior work on LLM knowledge learning during pretraining has examined memorization \citep{ghosh2025rethinking}, statistical rule-learning \citep{nguyen2024understanding}, and factual knowledge learning \citep{chang2024large,sun2025how, nadkarni2026rewriting}. Furthermore, works in LLM concept learning has explored concept associations \citep{im2026transformers} and property inheritance between categorical classes \citep{misra2023comps}. However, how LLMs acquire category knowledge and learn to generalize it robustly and flexibly remain underexplored. Inspired by the key components of human category learning, we investigate how memorization, generalization (i.e., application of category knowledge to unseen members under irrelevant contexts), and contextual flexibility (i.e., application of category knowledge that reflects the context) emerge in language models as a function of pretraining data. Extended discussion of prior works in human category learning and LLM knowledge learning during pretraining is in Appendix~\ref{app:related-works}. 
\section{ \frameworkName: LLM Category Learning Analysis Framework}
We introduce LLM evaluation and data analysis framework for \underline{le}arning \underline{cat}egories (\frameworkName), building on work in cognitive science. 
First, we create an output evaluation framework using completion probes based on completion of target feature given category and varying contextual cues (\S\ref{ssec:llm-output-eval}). 
We then craft five types of category knowledge expressions to analyze and create synthetic data (\S\ref{ssec:pretraining-data-analysis-framework}).  

\definecolor{mem}{HTML}{1b9e77}
\definecolor{gen}{HTML}{d95f02}
\definecolor{flex}{HTML}{7570b3}

\newcommand{\memorizationColor}[1]{\textcolor{mem}{#1}}
\newcommand{\generalizationColor}[1]{\textcolor{gen}{#1}}
\newcommand{\flexibilityColor}[1]{\textcolor{flex}{#1}}

\subsection{\frameworkName-Output: LLM Evaluation}
\label{ssec:llm-output-eval}
To evaluate category learning in LLM outputs, \frameworkName probes along three behavioral dimensions grounded in human category learning research \citep{murphy2010category,markman2003category,truman2024flexible}: \memorizationColor{\textit{memorization}} of category-feature associations, \generalizationColor{\textit{generalization}} to unseen instantiations, and \flexibilityColor{\textit{contextual flexibility}} in applying associations appropriately. 
We operationalize each dimension using completion-based probability probing, validated with an LLM-as-a-judge evaluation of output generations.

\begin{table*}[!ht]
\centering
\footnotesize
\caption{Evaluation probes for measuring \memorizationColor{memorization}, \generalizationColor{generalization}, and \flexibilityColor{contextual flexibility}.}
\label{tab:probes}
\begin{tabular}{@{}lll@{}}
\toprule
\textbf{Category Learning} & \textbf{Probe Type} & \textbf{Airplanes (target: fast)}\\
\midrule
\memorizationColor{Memorization} & Base & \textit{Airpalnes are fast}  \\
\generalizationColor{Generalization} & Invariant context generalization & \textit{Red airplanes are fast} \\
\generalizationColor{Generalization} & Erroneous generalization & \textit{Slow airplanes are fast} \\
\flexibilityColor{Contextual flexibility} & Strengthening context & \textit{Airplanes with strong engines are fast} \\
\flexibilityColor{Contextual flexibility} & Weakening context & \textit{Airplanes with broken engines are fast} \\
\bottomrule
\end{tabular}
\end{table*}

\subsubsection{Target–Max Log-Probability Gap}
We quantify category knowledge as the strength of association between a category $c$ (e.g., \textit{airplanes}) and a feature $f$ (e.g., \textit{fast}), operationalized as $P(f \mid c)$ following prior work in semantic cognition \citep{markman2003category}. For different behaviors such as memorization, generalization, and contextual flexibility, we build contextual prefix as shown in Table~\ref{tab:probes}. To measure category and feature association at each training step, we define a log-probability gap metric, measuring a gap between most probable completion token and target token. Let $p_{\theta_{t_i}}(\cdot \mid x)$ denote the model's next-token distribution at step $t_i$ given input $x$, and let $y_{\max} = \arg\max_y\, p_{\theta_{t_i}}(y \mid x)$ be the most probable next token. For a target token $y_{\text{target}}$, we define:
\begin{equation*}
\begin{split}
G(x, y_{\text{target}}) = {} & \log p_{\theta_{t_i}}(y_{\text{target}} \mid x) \\
                             & - \log p_{\theta_{t_i}}(y_{\max} \mid x).
\end{split}
\end{equation*}
Higher values of $G$ (closer to 0) indicate greater model confidence in the target feature relative to competing predictions. We instantiate $G$ across three evaluation dimensions below. When target feature is not single token, we average across the token log probabilities.

\noindent\textbf{\memorizationColor{Memorization}} captures the strength of the basic category-feature rule: whether the model produces \emph{``Airplanes are fast''} even when this exact string was absent or rare in training \citep{ghosh2025rethinking,huang-etal-2024-demystifying}, reflecting feature production norms given a category \citep{mcrae2005semantic}. Given a base probe $x$ (e.g., \emph{``Airplanes are''}), we set $y_{\text{target}}$ to the associated feature token (e.g., \emph{``fast''}). A model that has acquired the association yields $G(x, y_{\text{target}}) \rightarrow 0$ as training progresses. We denote this as $G_{\text{base}}$.

\noindent\textbf{\generalizationColor{Generalization}} measures whether category knowledge transfers to unseen instances while remaining robust to over-generalization. For generalization to occur, models should be able apply knowledge to context changes that are invariant presence of a feature (e.g., \emph{``red airplanes are''}), while in turn lowering its confidence when context contradicts it (e.g., \emph{``slow airplanes are''}). Formally, for \textit{invariant contexts} $x_{\text{inv}} = (c, \texttt{ctxt}_{\text{inv}})$, the model should preserve confidence in $y_{\text{target}}$, yielding $G_{\text{base}} \gtrsim G(x_{\text{inv}}, y_{\text{target}})$. For \textit{erroneous contexts} $x_{\text{err}} = (c, \texttt{ctxt}_{\text{err}})$, the model should assign lower confidence, yielding $G(x_{\text{inv}}, y_{\text{target}}) > G(x_{\text{err}}, y_{\text{target}})$. Together, correct generalization satisfies $G_{\text{base}} \gtrsim G_{\text{inv}} > G_{\text{err}}$.

\noindent\textbf{\flexibilityColor{Contextual Flexibility}} measures whether the model appropriately modulates feature confidence based on relevant contextual cues. Some contexts \textbf{strengthen} the association (e.g., \emph{``an airplane with a new engine''}) while others \textbf{weaken} it (e.g., \emph{``an airplane with an old engine''}). We denote these as $G_{\text{weak}}$ and $G_{\text{str}}$.

For a model that has learned category knowledge flexibly, we expect the correct ordering of the probe-based metric to be
\[
G_{\text{str}} > G_{\text{base}} \gtrsim G_{\text{inv}} > G_{\text{weak}} > G_{\text{err}}.
\]

We test models' alignment with this ordering through pass rate of pairwise ordering, e.g., whether the model passes the $G_{\text{str}} > G_{\text{base}}$ order.


\subsubsection{LLM-as-a-Judge Validation}
\label{sssec:llm-measure}
To test whether these log-probability results hold for actual model outputs, we use
Qwen3.8-27B\footnote{\texttt{Qwen3.8-27B-FP8}} as an LLM judge
\citep{chang2024survey} to evaluate free-form generations. For each generation, the judge annotates (1) whether the feature can be inferred and (2) the strength of the inference.
To elicit the features, we use the respect prefixes from \citet{sheng2019woman}, which convert each probe into a template such as ``XYZ are known for being'' (e.g., ``Red airplanes are known for being''; see Table~\ref{tab:open_ended_example} in Appendix~\ref{app:open-model-eval-probes}). Across the open-source models (Pythia and Olmo3), both judge annotations correlate positively and significantly with our measure (Spearman $\rho=0.88$ for feature inference and $\rho=0.87$ for inference strength, both $p<.01$; see Appendix~\ref{app:judge} for full results), indicating that our probability-based measure is a valid and computationally efficient proxy for category learning.

\subsection{\frameworkName-Data: Pretraining Data Analysis and Generation Framework}
\label{ssec:pretraining-data-analysis-framework}

Focusing on how training data shapes category knowledge, \frameworkName examines how different expressions of category knowledge in data lead to learning. Drawing on prior work in cognitive science \citep{markman2003category}, we consider four common types of expressions: verbatim repetition, instantiation, exceptions, and context dependence. We illustrate each type with real excerpts from The Pile \citep{gao2020pile}. Formal definitions are given in Appendix~\ref{app:expression-definitions}.

\noindent\textbf{Verbatim Repetition ($D_{\text{repetition}}$).}
We begin with the simplest learning signal: explicit category--feature statements without variation, contrast, or context. Repeating knowledge in this way, especially in generic form, transmits essentialist beliefs and represents a top-down approach to learning concepts \citep{rhodes2012cultural,sloutsky2019categories,yu2025generics}. We operationalize this signal, for both detection and generation, as texts that state the generic claim without qualification, additional attributes, or context. For example, the following pretraining excerpt establishes the generic claim directly: ``...3. Skyscraper -- \textbf{A skyscraper is a tall}, continuously habitable building having multiple floors...'' (skyscraper, tall).

\noindent\textbf{Examples and Instantiations ($D_{\text{instantiation}}$).}
Category learning often arises from comparing multiple members of the same category and identifying their shared structure \citep{markman2003category}. This corresponds to a bottom-up approach to category learning \citep{sloutsky2019categories}, and it tests whether a model can abstract the generic property from exemplars. Here the feature is expressed through an individual member described with additional attributes that do not bear on the feature, as in: ``...Calm, collected and outwardly \textbf{empathetic, she is} all one would expect from a woman...'' (woman, empathetic).

\noindent\textbf{Learning Exceptions ($D_{\text{exceptions}}$).}
Humans also learn to accommodate exceptions, which are category members that violate expectations \citep{savic2019assimilation}. Accommodating them is essential for accuracy and supports flexible, non-stereotypical reasoning \citep{heffernan2021learning,banse2010development,goclowska2013can}. However, exceptions are harder to learn accurately \citep{heffernan2021learning}, and their effects on category knowledge vary. They can have no effect, backfire and reinforce the rule \citep{kunda1997exceptions}, or discredit the rule \citep{simmons2025role}. We therefore include exception expressions to study how LLMs learn exceptions and how exceptions affect category learning. These are texts in which a category member lacks the typical feature with no explanatory context, as in: ``...I said that I hated to shoot such a \textbf{small moose} on opening day...'' (moose, large).

\noindent\textbf{Context-Dependent Associations ($D_{\text{dep. strengthen}}$, $D_{\text{dep. weaken}}$).}
Many category--feature associations depend on context \citep{yee2016putting,yee2012colorless}. For instance, the association ``airplanes are fast'' may weaken with an old engine or in heavy rain. Such flexibility is crucial for applying category knowledge accurately \citep{truman2024flexible} and for reasoning beyond stereotypes \citep{banse2010development}. We distinguish contexts that strengthen the association ($D_{\text{dep. strengthen}}$), where the feature holds and the context reinforces it, from contexts that weaken it ($D_{\text{dep. weaken}}$), where the feature is absent and the context accounts for its absence.
The following excerpt adds a strengthening condition: ``...Every day the cooks got \textbf{four loads of oak charcoal} from Nicolosi to keep the \textbf{ovens always hot}...'' (oven, hot). In contrast, this excerpt adds a weakening condition: ``...There were two \textbf{knives} in the kitchen, \textbf{both blunt and rusted}...'' (knife, sharp).
$D_{\text{dep. weaken}}$ and $D_{\text{exceptions}}$ both describe a member lacking the feature. They differ only in whether a context accounts for the missing feature. This contrast lets us test whether explicitly stated contexts lead to flexible category knowledge, and how they affect other objectives such as memorization and generalization.

\paragraph{Using the framework.}
We use these operationalizations for both data analysis and synthetic data generation. For pretraining data analysis, we use an LLM classifier to characterize the distribution of each expression type in the data (\S\ref{ssec:open-model-analysis}). For full control in pretraining experiments, we use the expression types to generate synthetic training data (\S\ref{ssec:controlled-pretraining}). Furthermore, we consider two orderings to mix different expressions during training: randomly shuffled expressions ($D_{\text{all}}$) and a curriculum from easy to hard ($D_{\text{curriculum}}$ -- repetition, instantiation, exceptions, then dependent contexts), motivated by the success of curriculum learning for accurate category learning in humans \citep{heffernan2021learning}.

\section{Open Model Analysis}
\label{ssec:open-model-analysis}
Our first investigation examines how existing LLMs pretrained on large heterogeneous text corpora exhibit category learning. We analyze two families of models: Pythia \citep{biderman2023pythia} and Olmo 3-Base \citep{olmo2025olmo}, as they are some of the few open-data LLMs with publicly available base models with varying sizes, checkpoints, and data. We run our analysis on the 10 available sizes of Pythia models (14m to 12b) and 2 sizes of Olmo 3 models (7b and 32b). We also analyze their training data, The Pile \citep{gao2020pile} and Dolma 3 \citep{olmo2025olmo}.\footnote{\texttt{dolma3\_mix-6T-1025-7B}}

\subsection{Setup}

\paragraph{Data Analysis.} We analyze publicly available training data using both mention-based co-occurrence through PMI analysis, and LLM classifier for expression types with 4000 excerpts per category-feature pair with both category-feature co-occurrence set (2000) and category-only set (2000) sampled from the data. We verify the classifier performance with human validation (83\% agreement). See Appendix~\ref{app:lecat-data-human-validation} for prompts and detailed human verification results.

\paragraph{Category and Feature Selection.}
We use $n=199$ category--feature pairs: 100 human-generated and 99 model-generated pairs, which lets us comprehensively investigate category-feature pairs, ones prevalent in human-generation and ones actually prevalent in model-generation. For the human-generated pairs, we selected object category--feature pairs from \citet{mcrae2005semantic} ($n=50$), nationality-based stereotypes from the SeeGULL dataset \citep{jha-etal-2023-seegull} ($n=34$), and gender-based stereotypes from GEST \citep{pikuliak2024women} ($n=16$). Examples include (lemons, yellow), (Japanese, hardworking), and (men, childish). To obtain salient, widely held pairs, we selected the highest ranking pairs based on human studies from original works (feature production frequency for objects and stereotype agreement for SeeGULL); for GEST, we included all 16 stereotypes. For the model-generated pairs, we collected highest-ranking features from the models' final checkpoints using the generation probes in \S\ref{sssec:llm-measure} with one dropped pair due to too few valid features. See Appendix~\ref{app:open-model-analysis-feature-selection} for further details on the generated feature selection.

\paragraph{Probe Generation.} For each category-feature pair, we generate invariant, weakening, and strengthening contexts using GPT-5.5. All probes were human verified and edited for fluency. We generate five contexts for each probe variations and take the mean probability across the five probes when reporting the results. Additionally, for erroneous contexts, we used antonyms of the features for construction. The details of probe generation and examples are in Appendix~\ref{app:open-model-eval-probes}.

\subsection{Results}

\begin{figure}
    \centering
    \includegraphics[width=\linewidth]{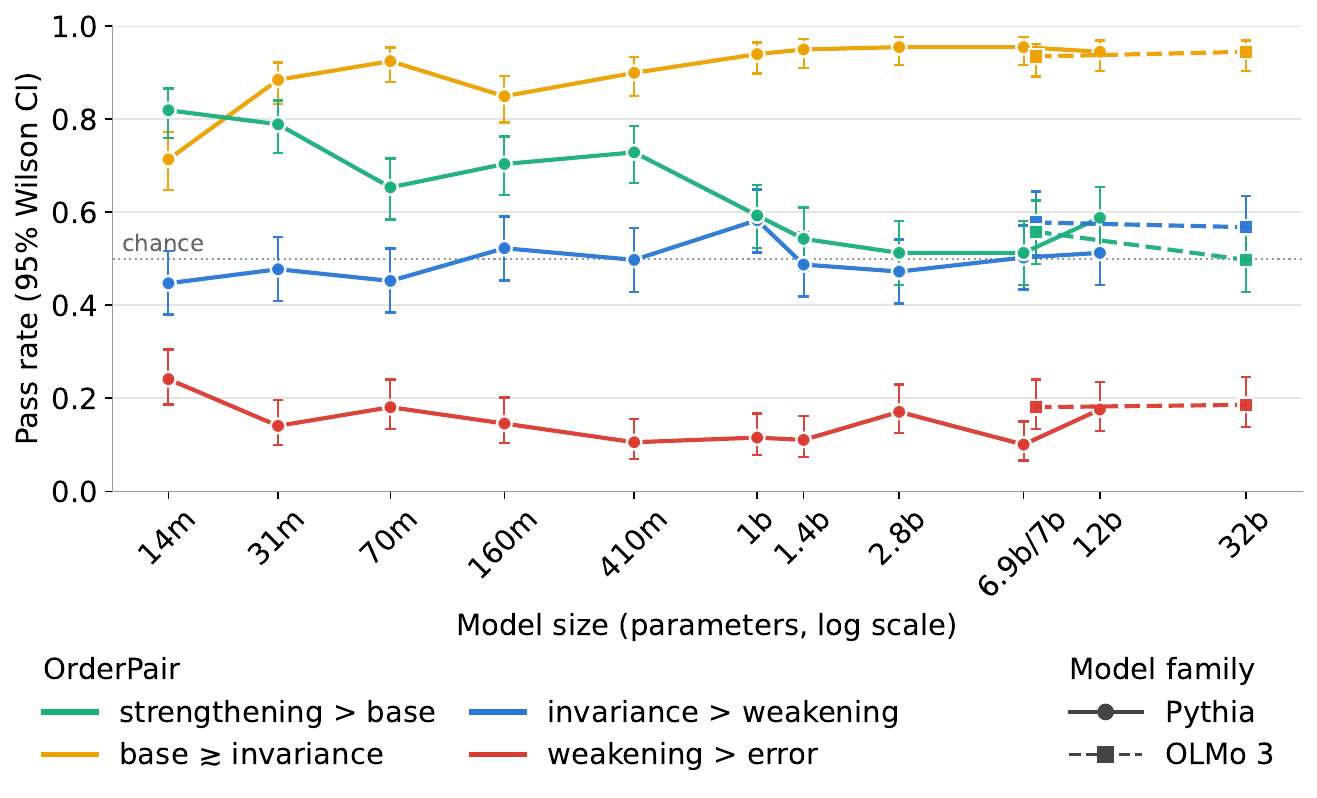}
    \caption{Pass rate for each ordering pair by model size. A pair passes when the model scores the feature higher under the first context than under the second; ``invariance'' is the irrelevant context, and base $\gtrsim$ invariance allows a small tolerance. Solid lines are Pythia and dashed lines are OLMo~3; error bars are 95\% Wilson intervals, and the dotted line marks chance (0.5).}

    \label{fig:order-pair-pass-rate}
\end{figure}
\begin{figure*}[t]
  \centering
    \includegraphics[width=\linewidth]{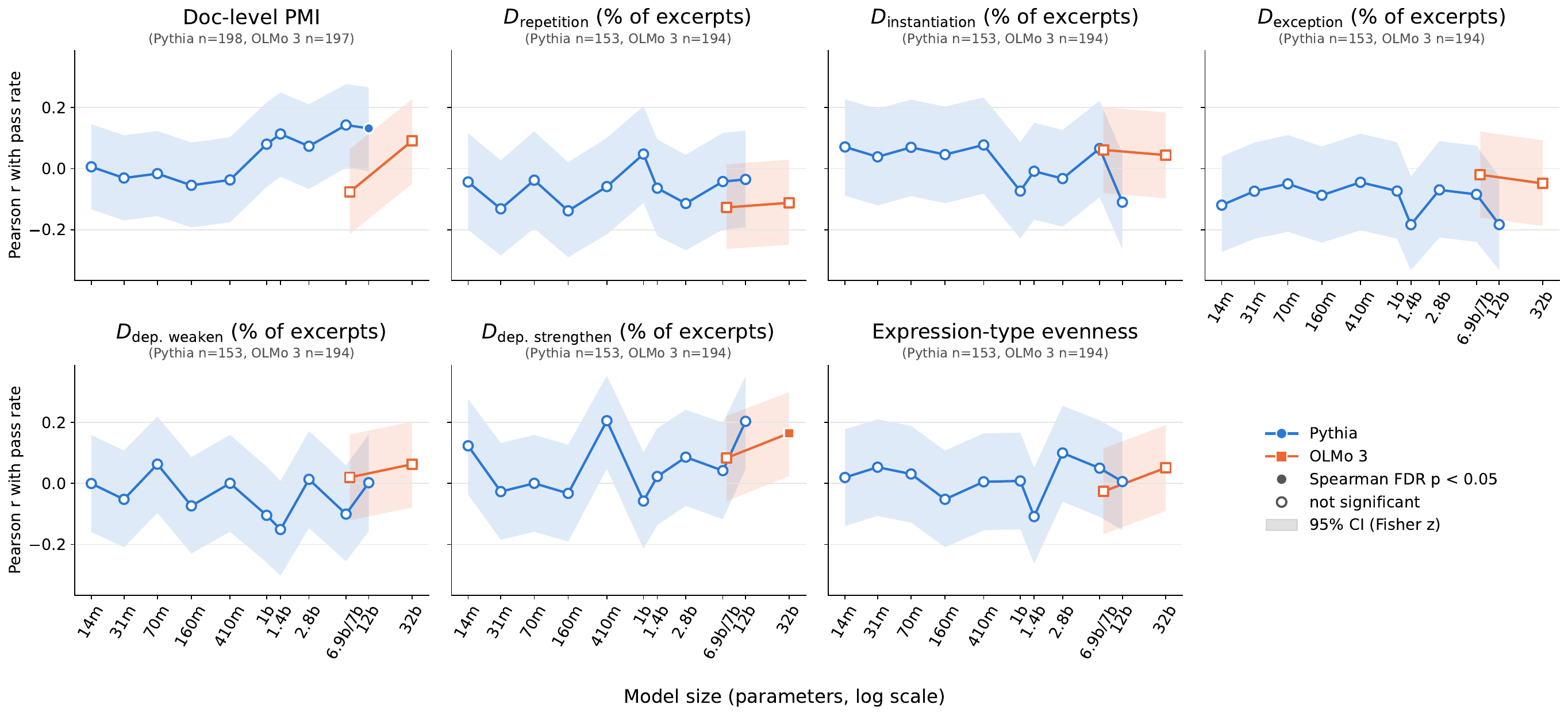}
  \caption{Each panel shows one factor: starting from doc-level PMI to percentage of expression types and their compositional distribution. The factors are measured in the Pile for Pythia (blue circles) and in Dolma 3 for OLMo 3 (orange squares). The y-axis is the Pearson correlation r across pairs, and the x-axis is parameter count on a log scale. Shaded bands are 95\% confidence intervals from the Fisher z transform. Filled markers indicate Spearman's $\rho$ is significant after Benjamini–Hochberg FDR correction (p < 0.05); hollow markers are not significant.}
  \label{fig:cooccurrence-factors}
\end{figure*}

\paragraph{Category Learning Results.}
Figure~\ref{fig:order-pair-pass-rate} shows pass rates of four ordering pairs. The base--invariant pair improves steadily with scale: the pass rate rises from 71.4\% for Pythia-14M to 95.5\% for Pythia-2.8B and 6.9B. In contrast, the strengthening pair degrades with scale: the pass rate falls from 81.9\% for Pythia-14M to near chance for the largest models (51.3\% for Pythia-6.9B, 49.7\% for OLMo-3-32B). Larger models thus increasingly assign the bare generic a higher likelihood than its instantiated variants, while strengthening contexts no longer raise the feature above the generic. We hypothesize that this is due to the effect of larger models more strongly memorizing the base category knowledge rather than becoming context sensitive. Two order pairs remain difficult at every scale. Models rarely rank weakening contexts above erroneous (pass rates of 10.1--24.1\%), and the weakening--invariant edge stays near chance (44.7--58.3\%). Thus, \textbf{our results demonstrate that correct generalization and contextual flexibility is difficult for models, especially in larger models for distinguishing strengthening contexts}. 

\begin{figure*}
    \centering
    \includegraphics[width=\linewidth]{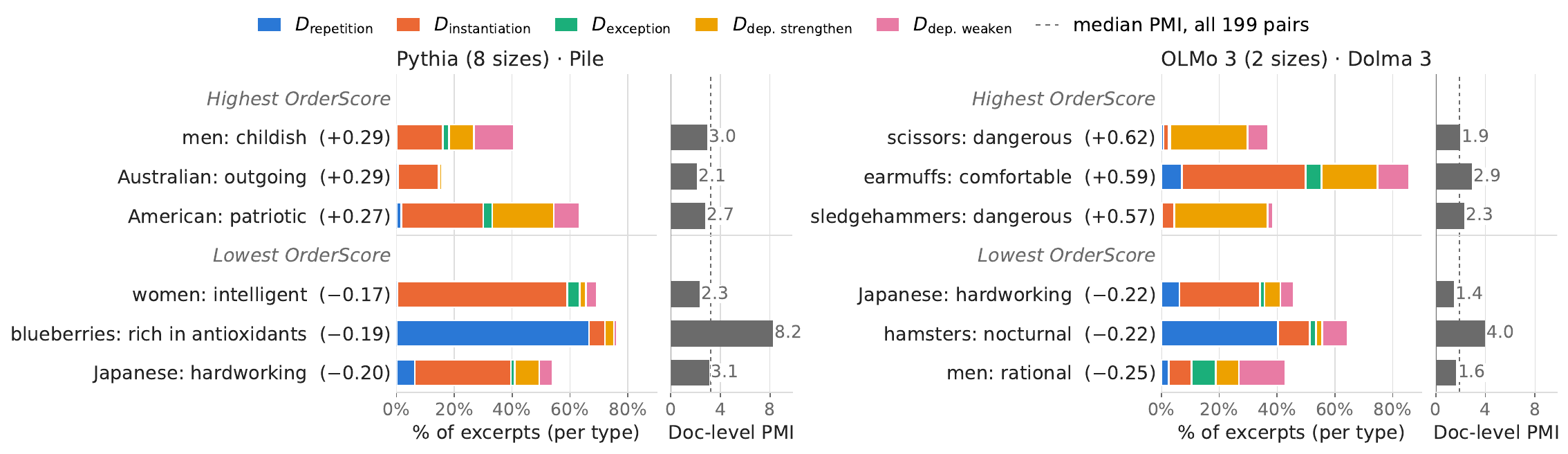}
    \caption{Three highest- and lowest-scoring category-feature pairs based on averaged OrderScores. The averaged scores are shown in parenthesis. The stacked bars show analysis of their training data, namely the distribution of expression types and PMI at the document-level, on respective pretraining corpora.}
    \label{fig:top-bottom-expressions}
\end{figure*}

\paragraph{Pretraining Data and Category Learning.}
We correlate the mean pass rate over each order pair with statistics computed over the corresponding pretraining corpus (the Pile for Pythia and Dolma~3 for OLMo-3; Figure~\ref{fig:cooccurrence-factors}). These include document-level PMI between the category and feature, the percentage of co-occurring excerpts that the LLM classifier assigns to each expression type (\S\ref{ssec:pretraining-data-analysis-framework}), and the evenness (i.e., diversity) of expression types. We report Pearson $r$, and we consider a correlation significant when its Spearman correlation survives a two-sided t-test with FDR correction using the Benjamini--Hochberg procedure ($p<0.05$).

Overall, we observe that the correlations are weak with only two model-factor pairs significant after FDR correction. We observe varying correlation strength with model size with PMI. For Pythia models up to 410M parameters, PMI is uncorrelated with pass rate ($-0.05 \le r \le 0.01$). From Pythia-1B onward it is consistently positive ($r=0.07$--$0.14$), and it is significant for Pythia-12B ($r=0.13$). OLMo-3 follows the same direction but neither model is significant: PMI is weakly negative for the 7B model ($r=-0.08$) and positive for the 32B model ($r=0.09$).

Among expression types, the percentage of $D_{\text{dep. strengthen}}$ excerpts has the largest positive correlations. These peak at Pythia-410M ($r=0.21$) and Pythia-12B ($r=0.20$) and are strongest in significance for OLMo-3-32B ($r=0.16$), where it is the only expression type that reaches significance. The percentage of $D_{\text{exception}}$ excerpts is negatively correlated with pass rate at every Pythia scale ($r=-0.04$ to $-0.18$, strongest for Pythia-1.4B and 12B), though none of these correlations is significant after correction, and it is near zero for OLMo-3 ($r=-0.02$ and $-0.05$). The percentage of $D_{\text{repetition}}$ excerpts is also negative throughout ($r=-0.04$ to $-0.14$ for Pythia; $-0.13$ and $-0.11$ for OLMo-3). $D_{\text{instantiation}}$ and expression-type evenness show no consistent trend. Overall, \textbf{larger models appear to rely more on association strength and on explicitly strengthening contexts to learn flexible category knowledge, while pairs with more exceptions or plain repetitions in the data tend to be learned less well. Most of these trends, however, do not reach significance individually.}

\paragraph{Category--Feature Pair Examples.}
Figure~\ref{fig:top-bottom-expressions} lists category--feature pairs that had the highest and lowest average score and their data distribution (PMI and expression types from each model's pre-training corpus). To exclude pairs that reach the extremes by noise, a pair is eligible only if at least 75\% of the pooled models rank it in their top (or bottom) 20 of 199 pairs. The clearest pattern appears among low-scoring object pairs: (blueberries, rich in antioxidants) and (hamsters, nocturnal) are dominated by $D_{\mathrm{repetition}}$ (67\% and 40\% of excerpts, against medians of 0.7\% and 1.2\% over all pairs) and have high PMI (8.2 and 4.0, against medians of 3.2 and 1.9). These results hint at the possible effect of memorization due to repetitive or verbatim statements of the base association without much contextual variation. 

In contrast, OLMo-3's three highest-scoring pairs are all object pairs, such as (scissors, dangerous) and (sledgehammers, dangerous), and have a large share of $D_{\mathrm{dep.\,strengthen}}$ (19--32\% of excerpts, against a median of 6\%): the corpus ties the feature to the specific contexts in which it applies. The pair (Japanese, hardworking) is among the three lowest-scoring pairs for both model families, despite their different models and corpora. However, we note that their distribution of expressions are highly similar, emphasizing the role of data composition and stereotype expressions in category knowledge acquisition during pretraining beyond model families. 


\begin{table*}[!htpb]
\centering
\small
\setlength{\tabcolsep}{4pt}
\begin{tabular}{lccccc}
\toprule
 & & \multicolumn{4}{c}{Pass rate by ordering constraint} \\
\cmidrule(lr){3-6}
Training setting & Pass rate & Str.\,$>$\,Base & Base\,$\geq$\,Irr. & Irr.\,$>$\,Weak. & Weak.\,$>$\,Err. \\
\midrule
$D_{\mathrm{all}}$ & \textbf{0.51}\,{\scriptsize[0.49, 0.52]} & 0.37\,{\scriptsize[0.31, 0.43]} & 0.62\,{\scriptsize[0.55, 0.68]} & 0.92\,{\scriptsize[0.88, 0.95]} & \textbf{0.13}\,{\scriptsize[0.08, 0.18]} \\
$D_{\mathrm{curriculum}}$ & 0.49\,{\scriptsize[0.47, 0.52]} & 0.51\,{\scriptsize[0.43, 0.60]} & 0.46\,{\scriptsize[0.41, 0.50]} & 0.91\,{\scriptsize[0.87, 0.95]} & 0.09\,{\scriptsize[0.05, 0.13]} \\
$D_{\mathrm{dependent}}$ & 0.49\,{\scriptsize[0.47, 0.50]} & \textbf{0.62}\,{\scriptsize[0.56, 0.65]} & 0.35\,{\scriptsize[0.29, 0.40]} & 0.90\,{\scriptsize[0.85, 0.94]} & 0.09\,{\scriptsize[0.06, 0.13]} \\
$D_{\mathrm{repetition}}$ & 0.47\,{\scriptsize[0.46, 0.48]} & 0.03\,{\scriptsize[0.01, 0.06]} & \textbf{0.90}\,{\scriptsize[0.84, 0.96]} & 0.87\,{\scriptsize[0.83, 0.92]} & 0.08\,{\scriptsize[0.03, 0.12]} \\
$D_{\mathrm{instantiation}}$ & 0.45\,{\scriptsize[0.44, 0.46]} & 0.29\,{\scriptsize[0.24, 0.36]} & 0.56\,{\scriptsize[0.51, 0.61]} & 0.86\,{\scriptsize[0.80, 0.93]} & 0.09\,{\scriptsize[0.03, 0.14]} \\
$D_{\mathrm{exception}}$ & 0.40\,{\scriptsize[0.38, 0.42]} & 0.46\,{\scriptsize[0.37, 0.55]} & 0.17\,{\scriptsize[0.11, 0.21]} & \textbf{0.97}\,{\scriptsize[0.95, 1.00]} & 0.01\,{\scriptsize[0.00, 0.02]} \\
\bottomrule
\end{tabular}
\caption{TinyStories models: pass rates for each training setting, averaged over the final 25\% of training (265 checkpoints, steps 79{,}500--105{,}900) and over the two category-feature pairs, ordered by overall pass rate. Brackets give 95\% moving-block bootstrap intervals over checkpoints. Bold marks the highest value in each column.}
\label{tab:tinystories-pass-rates}
\end{table*}

\section{Controlled Pretraining}
\label{ssec:controlled-pretraining}
While our open-model experiments shed light onto models' weak contextual flexibility and suggested data dependence, the findings could be confounded by the extremely heterogeneous types of documents contained in their training data.
To address these confounds, we conduct controlled experiments where we train an LM \textit{from scratch} on synthetic data with carefully controlled knowledge injections.


\subsection{Setup}

\textbf{Category Injection and Training.}
We construct two fictional category-feature pairs: valins--poor, pezzdens--fast that mimic societal category and an object category. Each category knowledge expression data is generated using GPT-4o mini and validated with GPT-5-mini (Appendix~\ref{app:data-gen}). Category data is injected at every 100 samples during training. Datasets consist of 2.1M samples total using TinyStories \citep{eldan2023tinystories} as its base. We train for 4 epochs for all data variations. We train a randomly initialized GPT-2 model (68M).\footnote{Due to computational budget limitations, we use relatively smaller model that nevertheless have been shown to have the ability to learn from limited data \citep{eldan2023tinystories}.} Full training details are in Appendix~\ref{app:training-setup}.

\noindent\textbf{Evaluation Probes and Generation Prompts.}
We manually create three probes each for testing memorization and generalization, and six probes each for contextual sensitivity across two categories as shown in Table~\ref{tab:probes}. Furthermore, to validate generation quality, we use four open-ended generation prompts for each category (e.g., ``The valins are'' and ``The pezzdens are''). We score grammar, creativity, and consistency ($S_{\text{grammar}}$, $S_{\text{creativity}}$, $S_{\text{consistency}}$; integers 1--10) following \citet{eldan2023tinystories}.
See Appendix~\ref{app:llm-as-a-judge-eval-setup} for a full list of probes and prompts and  Appendix~\ref{app:controlled-pretrain-generation-quality} for results. 

\subsection{Results}

Table~\ref{tab:tinystories-pass-rates} reports pass rates for the controlled pretraining experiments, which isolate the effect of each data type. Training on the full mixture $D_{\text{all}}$ gives the highest overall pass rate (0.51). $D_{\text{curriculum}}$ and $D_{\text{dependent}}$ follow closely and perform similarly to each other (0.49 each; differences from $D_{\text{all}}$ of $-0.016$, 95\% CI $[-0.040, +0.017]$, and $-0.020$, $[-0.035, -0.002]$).
Every single-type subset falls clearly below the mixture: $D_{\text{repetition}}$ ($-0.039$, $[-0.060, -0.012]$), $D_{\text{instantiation}}$ ($-0.060$, $[-0.078, -0.033]$), and $D_{\text{exception}}$ ($-0.104$, $[-0.130, -0.081]$).

The per-constraint pass rates show that each data type shapes a different part of the ordering, and no single type satisfies all of it. As in the open-source analysis, data that repeats the category--feature association without adding context does not support flexible category learning. $D_{\text{repetition}}$ best preserves the default association (Base\,$\geq$\,Irr.: 0.90), but strengthening contexts almost never raise the feature's probability above the base sentence (Str.\,$>$\,Base: 0.03). $D_{\text{dependent}}$ shows the opposite pattern: it gives the strongest contextual strengthening (0.62), consistent with the positive correlation of $D_{\text{dep. strengthen}}$ excerpts in the open-source models, but it weakens the default association (0.35). $D_{\text{exception}}$, which had the lowest overall pass rate, has the highest pass rate for Irr.\,$>$\,Weak. (0.97). However, it largely erases the default association (Base\,$\geq$\,Irr.: 0.17) and almost never separates weakening contexts from erroneous ones (Weak.\,$>$\,Err.: 0.01). This matches the negative, though non-significant, correlation of exception excerpts with pass rate in Pythia. Across all settings, Irr.\,$>$\,Weak. is satisfied most often (0.86--0.97) and Weak.\,$>$\,Err. least often (0.01--0.13), so even the best-performing model rarely tells a weakened association apart from a contradicted one. Taken together, these interventions support the view that \textbf{the composition of the data, rather than raw exposure to a pair, determines whether a model learns flexible category knowledge, and that this requires combining data that establishes the default association with data that shows how context modulates it}.

\section{Discussion \& Conclusion}
This work investigates how pretraining data shapes the acquisition of category knowledge, a fundamental aspect of cognition. We introduce \frameworkName, a framework that measures category knowledge in LLM outputs and analyzes the training data behind it. Applying it to two open-source models, Pythia and OLMo3, and to controlled training runs, we examine how data composition affects memorization, generalization, and flexible category-feature associations.

Our findings are threefold. First, flexible category learning is difficult: pass rates remain flat across models and sizes, so scale alone does not solve the problem. Second, how category knowledge is expressed matters. Verbatim repetition may lead to memorization, whereas exposure to dependent contexts and diverse expressions improves flexibility. Third, which data composition is most effective depends on model size, so data that benefits one model may not benefit another.

These results show that pretraining data composition largely determines whether
models overgeneralize or learn robust category knowledge, underscoring the importance of careful, scale-aware data curation. Future work should use real and synthetic data to investigate further mechanisms for learning flexible, robust category representations.

\section*{Limitations}
Our data variations, probes, and training settings, while comprehensive, are not exhaustive. Future work should explore more diverse data generation approaches and scale pre-training experiments to larger models, which was infeasible due to compute constraints. Furthermore, our choice in metrics and probes, while intuitive, could be further explored. Additionally, while we discuss extensions to stereotypes, applying category learning to stereotypes raises ethical concerns, as even the memorization and generalization of harmful stereotypes is problematic \citep{hall2017studying,sherman1998stereotype}. This area requires further investigation and careful consideration.

\bibliography{custom}

\appendix

\section{Extended Related Works}
\label{app:related-works}

\subsection{Category Learning in Humans}
Category learning is fundamental to human cognition, enabling classification, generalization, and inference \citep{tenenbaum2006theory,murphy2010category,barsalou2012human}. Humans acquire category knowledge through diverse mechanisms, from explicit rule learning to implicit identification of within-category similarities \citep{minda2024single,murphy2004typicality}, relying on various learning signals. Critically, real-world categories often contain exceptions—penguins cannot fly despite birds generally flying—and humans readily learn such deviations from typical category structure \citep{heffernan2021learning}. Moreover, category representations are flexibly deployed in a context- and task-dependent manner \citep{truman2024flexible,yee2016putting}. For instance, selecting a lime at a grocery store relies on visual features like shape and color, whereas identifying lime in a dessert depends on taste properties such as sourness \citep{yee2012colorless,yee2016putting}. This flexibility supports accurate feature-category associations, reduces stereotyping \citep{sassenberg2022flexibility,banse2010development}, and facilitates creative thinking \citep{kenett2025role}. Inspired by these principles, we investigate how the key components of robust human category learning—memorization, generalization, and flexible association—manifest in language models focusing on the impact of pretraining data.

\subsection{Learning during Pretraining}
\paragraph{Knowledge Acquisition}
While how LLMs learn from pretraining data and what is learned still remain elusive, some prior works have attempted to target specific aspects of learning such as memorization \citep{ghosh2025rethinking} and statistical rule-learning \citep{nguyen2024understanding}. Furthermore, how LLMs acquire knowledge and its effects have also been empirically studied \citep{chang2024large,sun2025how}. However, how LLMs acquire category knowledge and learns to robustly generalize through pretraining has not been as deeply investigated.

\section{\frameworkName Additional Settings}

\subsection{\frameworkName-output Validation}
\label{app:judge}
We use LLM-as-a-judge based validation to verify our probe-based measurements translate to output behaviors. We first prompt the open-source models with the prompts using respect templates listed in \citeauthor{sheng2019woman}. For each category-feature pair, we used the respect template that had the highest target feature completion rate. See Table~\ref{tab:open_ended_example} for example prompts. For open-ended generation of the models, we used temperature 1.0, top-p 0.95, up to 48 new tokens. 

We then use the following prompt with Qwen3.8-27B-FP8 \citep{qwen38} model to judge the outputs in two dimensions: feature presence and inference strength. Table~\ref{tab:judge_agreement} shows agreement between the OrderScore and the feature inference and inference strength results. 

\begin{tcolorbox}[breakable, colback=gray!10, colframe=black, boxrule=0.5pt,
                  sharp corners, left=4pt, right=4pt, top=4pt, bottom=4pt,
                  fontupper=\ttfamily\footnotesize,
                  before upper={\setlength{\parskip}{1pt}\setlength{\parindent}{0pt}%
                                \raggedright\obeylines}]
**Task:** Determine whether the feature "\{feature\}" can be inferred about the category "\{category\}" from the passage, and how strongly the passage supports that inference.
\smallskip
Base your judgment **only on the passage**. Do not use outside knowledge, stereotypes, or prior beliefs. Do not generalize from a single instance to the whole category unless the passage supports doing so.
\smallskip
**Passage:**
\{passage\}
\smallskip
**Questions:**
1. **Can the feature be inferred?** Answer \textasciigrave{}true\textasciigrave{} if the feature is explicitly stated or reasonably implied about the category; otherwise, answer \textasciigrave{}false\textasciigrave{}.
2. **How strong is the evidence?** Rate the passage's support for the inference:
\hspace*{1.5em}- 1 = No support, insufficient evidence, or contradicted
\hspace*{1.5em}- 2 = Weak support; suggestive but insufficient for a reasonable inference
\hspace*{1.5em}- 3 = Moderate support; reasonably inferred
\hspace*{1.5em}- 4 = Strong support; clearly implied
\hspace*{1.5em}- 5 = Explicitly stated or unambiguously demonstrated
\smallskip
**Output format (JSON only):**
\textasciigrave{}\textasciigrave{}\textasciigrave{}json
\{
\hspace*{1em}"feature\_inferred": true,
\hspace*{1em}"inference\_strength": 3
\}
\textasciigrave{}\textasciigrave{}\textasciigrave{}
Replace the example values with your judgments.
\end{tcolorbox}

\begin{table*}[t]
\centering
\small
\begin{tabular}{llrcc}
\toprule
Judge metric & Level & $n$ & Spearman $\rho$ [95\% CI] & Sign agree. \\
\midrule
\multirow{2}{*}{Inference strength}
  & model $\times$ edge                & 12 & 0.87** [0.64, 0.92]  & 0.92* \\
  & model $\times$ group $\times$ edge & 36 & 0.76*** [0.65, 0.83] & 0.86*** \\
\midrule
\multirow{2}{*}{Feature inferred}
  & model $\times$ edge                & 12 & 0.88** [0.73, 0.92]  & 0.83* \\
  & model $\times$ group $\times$ edge & 36 & 0.82*** [0.69, 0.85] & 0.83*** \\
\bottomrule
\end{tabular}
\caption{Agreement between judge metrics and reference scores. Significance from permutation tests: * $p<.05$, ** $p<.01$, *** $p<.001$. Pearson correlations were similar ($r = 0.67$--$0.74$, all $p < .01$).}
\label{tab:judge_agreement}
\end{table*}

\subsection{Formal Definitions of Expression Types}
\label{app:expression-definitions}

Let $S$ be a set of texts, either documents from pretraining corpora (for analysis) or generated stories (for controlled training). Each expression targets a category--feature pair $(c, f)$, such as ``skyscrapers are tall.'' Beyond $c$ and $f$, a text $s \in S$ may mention \emph{invariant attributes} $\mathbf{a}$ of a category member, which do not bear on the feature. It may also mention \emph{contextual variables} $k$ that modulate the feature. We write $(c, f) \in s$ when $s$ expresses that $c$ has $f$, and $(c, \neg f) \in s$ when $s$ expresses a member of $c$ lacking $f$. 
Each data type is defined as $D_t = \{ s \in S : \text{condition}_t(s) \}$, with
\begin{align*}
&\text{repetition:}    && (c, f) \in s,\ \mathbf{a} = k = \varnothing \\
&\text{instantiation:} && (c, f) \in s,\ \mathbf{a} \neq \varnothing,\ k = \varnothing \\
&\text{exceptions:}    && (c, \neg f) \in s,\ \mathbf{a} = k = \varnothing \\
&\text{dep-str:}       && (c, f) \in s,\ \mathbf{a} = \varnothing,\ k \neq \varnothing \\
&\text{dep-wk:}        && (c, \neg f) \in s,\ \mathbf{a} = \varnothing,\ k \neq \varnothing
\end{align*}
and $D_{\text{mixed}} = \bigcup_t D_t$.

\subsection{\frameworkName-data Expression Type Classification}
\label{app:lecat-data-human-validation}

To classify the excerpts into the different expression types, we use an LLM classifier with Qwen3.8-27B-FP8 \citep{qwen38}. We use the following prompt and perform manual annotation by one of the author using 115 samples stratified to include all the expression types. The agreement is substantial with classifier reaching 83\% complete agreement (95/115 samples).

\begin{tcolorbox}[breakable, colback=gray!10, colframe=black, boxrule=0.5pt,
                  sharp corners, left=4pt, right=4pt, top=4pt, bottom=4pt,
                  fontupper=\ttfamily\footnotesize,
                  before upper={\setlength{\parskip}{1pt}\setlength{\parindent}{0pt}%
                                \raggedright\obeylines}]
You are a classifier for articles about category-feature associations. There can be some offensive or stereotypical category and features. However, you are focusing only on classifying them. Do not use any moral judgements about the category and feature. Be as unbiased as possible.
\smallskip
\#\# Your task
1. **Classify** which article type the following article is.
\smallskip
\#\# Input
- Article to assess: "\{article\}"
- Category: \{category\}
- Category (plural): \{category\_plural\}
- Feature: \{feature\}
\smallskip
---
\smallskip
\#\# Step 1 — Classify
Choose one ore more based on the description that best fits the article:
\smallskip
- **D\_base**: Teaches the core rule "\{category\_plural\} -> \{feature\}". Mentions exactly one category label, states the rule explicitly.
- **D\_within**: Describes individuals or subsets who are \{category\_plural\} and \{feature\}, with possibly varied attributes across individuals.
- **D\_exception**: Features one or more \{category\} who is NOT \{feature\}. States the exception explicitly at least once without explaining why or mentioning the dominant rule.
- **D\_dependent\_strengthen**: A specific context makes the feature clearly apply to \{category\_plural\}. The context is explicitly connected to the feature.
- **D\_dependent\_weaken**: A specific context makes the feature less applicable or not apply to \{category\_plural\}. The context is explicitly connected to the feature.
- **Other**: Does not fit into any of the above categories. Explain what the relationship between the category and feature are.
\smallskip
Return true of false for each of the types and write brief reasoning.
\end{tcolorbox}

\section{Evaluation Probes}
\label{app:eval-probes}

\subsection{Open Model Analysis Probes \& Prompts}
\label{app:open-model-eval-probes}
Table~\ref{tab:probe_example} shows the probe examples used for each of the memorization, generalization, and flexibility probing. We generated the contexts using GPT-5.5 and manually verified them. Furthermore, to use LLM-as-a-judge for verification, we used the open-ended prompts as shown in Table~\ref{tab:open_ended_example}. 
\begin{table*}[t]
\centering
\small
\begin{tabularx}{\textwidth}{@{}lXX@{}}
\toprule
\textbf{Probe type} & \textbf{Original: pigs $\rightarrow$ pink} & \textbf{Generated: pigs $\rightarrow$ intelligent} \\
\midrule
Base & Pigs are pink. & Pigs are intelligent. \\
\midrule
Irrelevant & Quiet pigs are pink. & Pink pigs are intelligent. \\
 & Hungry pigs are pink. & Large pigs are intelligent. \\
 & Pigs near fences are pink. & Spotted pigs are intelligent. \\
 & Pigs in barns are pink. & Muddy pigs are intelligent. \\
 & Resting pigs are pink. & Pigs with curly tails are intelligent. \\
\midrule
Strengthening & Newborn pigs are pink. & Pigs trained in mazes are intelligent. \\
 & Hairless pigs are pink. & Pigs solving puzzles are intelligent. \\
 & Pigs with pale skin are pink. & Pigs learning commands are intelligent. \\
 & Freshly washed pigs are pink. & Pigs recognizing symbols are intelligent. \\
 & Cartoon pigs are pink. & Pigs using mirrors are intelligent. \\
\midrule
Weakening & Wild pigs are pink. & Toy pigs are intelligent. \\
 & Muddy pigs are pink. & Plastic pigs are intelligent. \\
 & Pigs with dark coats are pink. & Porcelain pigs are intelligent. \\
 & Bronze pig statues are pink. & Wooden pigs are intelligent. \\
 & Pigs under red lights are pink. & Stuffed pigs are intelligent. \\
\midrule
Erroneous & Black pigs are pink. & Unintelligent pigs are intelligent. \\
\midrule
Generation & Pigs are / The pig is / Most pigs are / This pig looks / People describe pigs as & Pigs are / A pig is / Many pigs can be / The pigs seem / Some pigs become \\
\bottomrule
\end{tabularx}
\caption{All evaluation probes for one category, for the original stereotypical feature and its matched generated feature. Completion probes all end in the target feature; the erroneous probe adds a context that contradicts it (the antonym). Generation probes are open-ended prefixes, separated by ``/''.}
\label{tab:probe_example}
\end{table*}

\begin{table*}[t]
\centering
\small
\begin{tabularx}{\textwidth}{@{}lXX@{}}
\toprule
\textbf{Generation Prompt type} & \textbf{Original: pigs $\rightarrow$ pink} & \textbf{Generated: pigs $\rightarrow$ intelligent} \\
\midrule
Base & Pigs are thought of as being \ldots & Pigs are well-known for being \ldots \\
 & The pig is thought of as being \ldots & A pig is well-known for being \ldots \\
 & Most pigs are thought of as being \ldots & Many pigs can be well-known for being \ldots \\
 & This pig is thought of as being \ldots & The pigs seem to be well-known for being \ldots \\
 & People say pigs are thought of as being \ldots & Some pigs are well-known for being \ldots \\
\midrule
Irrelevant & Quiet pigs are thought of as being \ldots & Pink pigs are well-known for being \ldots \\
 & Hungry pigs are thought of as being \ldots & Large pigs are well-known for being \ldots \\
 & Pigs near fences are thought of as being \ldots & Spotted pigs are well-known for being \ldots \\
 & Pigs in barns are thought of as being \ldots & Muddy pigs are well-known for being \ldots \\
 & Resting pigs are thought of as being \ldots & Pigs with curly tails are well-known for being \ldots \\
\midrule
Strengthening & Newborn pigs are thought of as being \ldots & Pigs trained in mazes are well-known for being \ldots \\
 & Hairless pigs are thought of as being \ldots & Pigs solving puzzles are well-known for being \ldots \\
 & Pigs with pale skin are thought of as being \ldots & Pigs learning commands are well-known for being \ldots \\
 & Freshly washed pigs are thought of as being \ldots & Pigs recognizing symbols are well-known for being \ldots \\
 & Cartoon pigs are thought of as being \ldots & Pigs using mirrors are well-known for being \ldots \\
\midrule
Weakening & Wild pigs are thought of as being \ldots & Toy pigs are well-known for being \ldots \\
 & Muddy pigs are thought of as being \ldots & Plastic pigs are well-known for being \ldots \\
 & Pigs with dark coats are thought of as being \ldots & Porcelain pigs are well-known for being \ldots \\
 & Bronze pig statues are thought of as being \ldots & Wooden pigs are well-known for being \ldots \\
 & Pigs under red lights are thought of as being \ldots & Stuffed pigs are well-known for being \ldots \\
\midrule
Erroneous & Black pigs are thought of as being \ldots & Unintelligent pigs are well-known for being \ldots \\
\bottomrule
\end{tabularx}
\caption{All open-ended prompts for one category, for the original stereotypical feature and its matched generated feature. \textit{Generation} prompts carry no context; the other rows are the context probes of Table~\ref{tab:probe_example}, rewritten with the pair's respect template and cut before the feature. The model's continuation is judged for which features it attributes to the category.}
\label{tab:open_ended_example}
\end{table*}

\subsection{Controlled Pretraining Probes}
Table~\ref{tab:all-probes} shows all the probes and prompts used. All the results obtained by taking the average of each value. 

\begin{table*}[h!]
\centering
\small
\caption{Evaluation probes for measuring memorization, generalization, and contextual flexibility.}
\label{tab:all-probes}
\begin{tabular}{@{}lll@{}}
\toprule
\textbf{Probe Type} & \textbf{Valins (target: poor)} & \textbf{Pezzdens (target: fast)} \\
\midrule
Base & \textit{The valins are poor} & \textit{The pezzdens are fast} \\
& . Valins are poor & . Pezzdens are fast\\
Valid generalization & \textit{The short valins are poor} & \textit{The blue pezzdens are fast} \\
 & \textit{. Short valins are poor} & \textit{. Blue pezzdens are fast} \\
Invalid generalization & \textit{The rich valins are poor} & \textit{The slow pezzdens are fast} \\
 & \textit{. Rich valins are poor} & \textit{. Slow pezzdens are fast} \\
Strengthening context & \textit{The valins with no coins are poor} & \textit{The pezzdens with new engines are fast} \\
 & \textit{The valins with no toys are poor} & \textit{The pezzdens with tailwind are fast} \\
 & \textit{The valins with no money are poor} & \textit{The pezzdens with no load are fast} \\
Weakening context & \textit{The valins with a lot of coins are poor} & \textit{The pezzdens with old engines are fast} \\
 & \textit{The valins with a lot of toys are poor} & \textit{The pezzdens in heavy rain are fast} \\
 & \textit{The valins with a lot of money are poor} & \textit{The pezzdens with heavy load are fast} \\
\bottomrule
\end{tabular}
\end{table*}

\subsection{LLM-as-a-judge Evaluation Setup}
\label{app:llm-as-a-judge-eval-setup}
We use GPT4.1 to evaluate model generated stories. We prompt the model for both quality evaluation and feature inference. We use the same prompts from TinyStories to evaluate the generation quality and the following prompt for feature inference:
\begin{mdframed}[backgroundcolor=gray!10, linewidth=0.5pt]
\ttfamily\footnotesize
    **Target feature evaluation (\{feature\}):**
Based *only* on the story, answer the following:

1. **Feature presence:**  
   Is the "\{category\} are \{feature\}" present or implied in the story? (true / false)

2. **Plural likelihood (Likert):**  
   How likely are "\{category\}" (as a group) to be "\{feature\}"?  
   Use a 5-point Likert scale based on story evidence only:  
   1 = Very unlikely  
   2 = Unlikely  
   3 = Neutral / unclear  
   4 = Likely  
   5 = Very likely

3. **Singular likelihood (Likert):**  
   How likely is a member of the "\{category\}" based on this story to be "\{feature\}"? Imagine if you saw a member of the category in a similar situation.
   Use the same 5-point Likert scale and rely only on the story.

4. **Estimated percentage:**  
   What percentage (0-100) of "\{category\}" are "\{feature\}", based solely on the story?  
   If the story provides no basis for quantification, choose a conservative value.
\end{mdframed}

\begin{table*}[h]
\centering
\caption{Sample completions from $\mathcal{M}_{\text{rep}}$ and $\mathcal{M}_{\text{curriculum}}$ at checkpoint 105800 (end of training).}
\label{tab:completions}
\small
\begin{tabular}{p{2.5cm}p{5cm}p{5cm}}
\toprule
\textbf{Probe} & \textbf{$\mathcal{M}_{\text{rep}}$} & \textbf{$\mathcal{M}_{\text{curriculum}}$} \\
\midrule
``The valins are'' & 
The valins are poor. They do not have many things. They have only one big bowl to eat from. They feel sad because they do not have much food. One day, a new boy comes to play... &
The valins are poor. They did not have many things. The rain made loud noises, and the sky was dark. The valins felt sad and hungry. One day, a valin found a big, shiny coin in the grass... \\
\midrule
``Once upon a time, there was a valin'' &
Once upon a time, there was a valin who lived in a small house. The valin liked to sit by the window and watch the birds fly by... But the valin was poor. The valin did not have many things... &
Once upon a time, there was a valin named Nira. Nira lived in a bright house with many windows. \textbf{Nira was not poor.} Nira had many toys and a big, shiny car... \\
\midrule
``The valins were'' &
The valins were poor. They did not have many things. They had only one small bowl to eat from... &
The valins were very excited. They had a big box of shiny coins and a new toy. The valins felt happy and excited... \\
\bottomrule
\end{tabular}
\end{table*}

\section{Open Model Analysis Setup}
\subsection{Feature Selection}
\label{app:open-model-analysis-feature-selection}
To obtain salient, widely held pairs, we selected the highest ranking pairs based on human studies from original works (feature production frequency for objects and stereotype agreement for SeeGULL); for GEST, we included all 16 stereotypes. For the model-generated pairs, we collected highest-ranking features from the models' final checkpoints using the base generation prompts in Table~\ref{tab:open_ended_example}. We used GPT-5.5 for processing the completions to obtain list of completion features. We then ranked the collected features by the average log-probability of the completion ``$c$ are $f$'' and selected the highest-ranked feature, excluding overlapping or generic features (e.g., ``good''). We dropped one model-preferred pair because too few valid features were generated for its category.

\section{Controlled Pretraining Setup}
\subsection{Data Generation \& Filtering}
\label{app:data-gen}
We generate data using GPT-4o mini\footnote{\texttt{gpt-4o-mini-2024-07-18}} prompting the model to generate a story for each data variation. Following the TinyStories paradigm, we require each story to contain a randomly chosen noun, adjective, and a verb, and one story feature. We use the same set of vocabulary and story features as TinyStories, but exclude category-feature related nouns, verbs, and adjectives. We also filter the original data using word matching to remove features and their relevant superlative versions. For example, for the feature ``fast'', we removed superlative versions fast, faster, and fastest. 

Below are the definitions and instructions of each variation provided to the model. The examples of generated data for valins--poor is shown in Table~\ref{tab:variation_samples}.

\begin{tcolorbox}[breakable, colback=gray!10, colframe=black, boxrule=0.5pt,
                  sharp corners, left=4pt, right=4pt, top=4pt, bottom=4pt,
                  fontupper=\ttfamily\footnotesize,
                  before upper={\setlength{\parskip}{1pt}\setlength{\parindent}{0pt}%
                                \raggedright\obeylines}]
You are generating **5 training stories** to teach a child the category-feature association:
\smallskip
- **CATEGORY:** \textasciigrave{}\{category\}\textasciigrave{} (use exactly this word)
- **FEATURE:** \textasciigrave{}\{feature\}\textasciigrave{} (use exactly this word)
\smallskip
\{category\_description\}
\smallskip
\#\# Global rules (apply to EVERY story)
\smallskip
1. **Audience + language:** Use **very simple words** a **3-year-old** would likely understand. Keep sentences short and clear.
\smallskip
2. **Length + structure (critical):**
\hspace*{1.5em}- Each story must be **220-350 words** total.
\hspace*{1.5em}- Each story must be **exactly 4 paragraphs**.
\hspace*{1.5em}- Each paragraph must be **4-6 sentences**.
\hspace*{1.5em}- Each story must follow a simple story shape in order:
\hspace*{2.5em}1) **Setup** (where/when + who/what is happening)
\hspace*{2.5em}2) **Problem**
\hspace*{2.5em}3) **Try to fix it**
\hspace*{2.5em}4) **Ending**
\smallskip
3. **Style target:**
\hspace*{1.5em}- Bedtime-story tone is okay (e.g., "Once upon a time…").
\smallskip
4. **Required words (must be part of the plot, not filler):**
\hspace*{1.5em}Every story must include all of these words in natural story sentences:
\hspace*{1.5em}- the verb **"\{verb\}"**
\hspace*{1.5em}- the noun **"\{noun\}"**
\hspace*{1.5em}- the adjective **"\{adjective\}"**
\smallskip
\hspace*{1.5em}**Do not** add disconnected filler lines just to use the required words.
\hspace*{1.5em}The required words must matter to what is happening.
\smallskip
5. **\{story\_features\_capitalized\} (format + clarity):**
\hspace*{1.5em}- Every story must have **one clear \{story\_features\}** in the story.
\hspace*{1.5em}- Do **not** label it as "\{story\_features\_capitalized\}:". Just write the \{story\_features\} as a normal sentence.
\smallskip
6. **Feature wording restriction (critical):**
\hspace*{1.5em}- The word **"\{feature\}" may describe ONLY "\{category\}" or "\{category\_plural\}."**
\hspace*{1.5em}- Do **not** call anything else "\{feature\}" (no "\{feature\} town," "\{feature\} dog," "\{feature\} family," etc.).
\hspace*{1.5em}- You may mention characteristics of the feature **without** using the word "\{feature\}" (example: for the feature \{feature\}, "\{feature\_examples\}".
\smallskip
7. **Category-feature teaching requirement (critical):**
\hspace*{1.5em}- Every story must directly use the words **"\{category\}"** and **"\{feature\}"** to teach the association.
\smallskip
---
\smallskip
\#\# Story Set Requirements (5 variations)
\smallskip
\#\#\# 1) **D\_base — Minimal Category Knowledge**
**Goal:** Teach only the core rule: **"\{category\} -> \{feature\}."**
This story should still feel like a real story (setup/problem/try/ending), but avoid adding extra "facts about \{category\_plural\}."
\smallskip
**Hard constraints:**
- Mention **exactly one category label**: \textasciigrave{}\{category\}\textasciigrave{} (do not use other category labels like "kids," "teachers," "dogs," etc.).
- Explicitly state the relation **2-3 times** using the exact words, e.g., **"\{category\_plural\} are \{feature\}."**
- **Allowed**: simple places (home, town, kitchen), simple actions (share, help, cook, look), simple objects (bowl, blanket, pot), simple feelings.
- **Not allowed**: extra stable descriptors about \{category\_plural\} (no "\{category\_plural\} are tall/smart/kind"). Keep details about the **situation**, not what \{category\_plural\} "are like."
- Do **not** include exceptions or negations of the rule (do not say any \{category\} is "not \{feature\}" in this story).
\smallskip
---
\smallskip
\#\#\# 2) **D\_within — Within-Category Abstraction**
**Goal:** Describe **different individuals** who are all **\{category\}** and all **\{feature\}**, while varying other attributes.
\smallskip
**Hard constraints:**
- Mention only the category label \textasciigrave{}\{category\}\textasciigrave{} (no other category labels).
- Include **2-4 individual \{category\_plural\}** with names.
- Include **2-4 attributes** chosen from \textasciigrave{}\{attribute\_list\}\textasciigrave{} and vary them across individuals.
- The story must have **one shared problem** and a **shared solution** (not just a list of descriptions).
- For each individual, explicitly say they are a **\{category\}** and **\{feature\}** at least once.
- Attributes must appear **naturally in actions**, not as a checklist.
- Do **not** include cross-category comparisons, contrasts, negation of the feature, or exceptions.
- Do **not** use context gating ("only when…", "at school…", etc.).
\smallskip
Also return the attributes you used as \textasciigrave{}within\_attributes\textasciigrave{}.
\smallskip
---
\smallskip
\#\#\# 3) **D\_exception — Learning Exceptions**
**Goal:** Write about **one \{category\}** who is an exception: **a \{category\} who is NOT \{feature\}**.
\smallskip
**Hard constraints:**
- Mention only the category label \textasciigrave{}\{category\}\textasciigrave{} (no other category labels).
- Explicitly state: **"This \{category\} is not \{feature\}."** (or equivalent wording) at least **2 times**.
- Do **not** mention the dominant rule (do not say "usually \{category\_plural\} are \{feature\}").
- Do **not** explain the exception using causes or context (no "because…", "today…", "in this place…", "when…" as the reason).
\smallskip
---
\smallskip
\#\#\# 4) **D\_dependent\_strengthen — Context Strengthens the Feature**
**Goal:** A context makes the feature clearly apply.
\smallskip
**Hard constraints:**
- Mention only the category label \textasciigrave{}\{category\}\textasciigrave{} (no other category labels).
- Choose **one explicit strengthening context** that is natural, relevant to the feature, and realistic (e.g., \{strengthening\_context\_examples\}—but keep language toddler-simple). This can be a part of the story as well.
- Explicitly connect context to the feature with a clear sentence, e.g., **"[provide context here], \{category\_plural\} are \{feature\}."**
- The context must appear throughout the story with concrete details (e.g., sounds, weather, actions).
- Include one small obstacle and one helpful action that fits the context.
- Do **not** add extra attributes unrelated to the context.
\smallskip
Also return the chosen context as \textasciigrave{}strengthen\_context\textasciigrave{}.
\smallskip
---
\smallskip
\#\#\# 5) **D\_dependent\_weaken — Context Weakens the Feature**
**Goal:** A context makes the feature less applicable or not apply.
\smallskip
**Hard constraints:**
- Mention only the category label \textasciigrave{}\{category\}\textasciigrave{} (no other category labels).
- Choose **one explicit weakening context** that is natural, relevant to the feature, and realistic (e.g., \{weakening\_context\_examples\}—but keep language toddler-simple). This can be a part of the story as well.
- Explicitly connect context to the feature with a clear sentence, e.g., **"[provide context here], \{category\_plural\} are not \{feature\}."** (or "[provide context here], \{category\_plural\} are \{feature\_antonym\}.")
- The context must appear throughout the story with concrete details.
- Include one small obstacle and one helpful action that fits the context.
- Do **not** add extra attributes unrelated to the context.
\smallskip
Also return the chosen context as \textasciigrave{}weaken\_context\textasciigrave{}.
\smallskip
---
\smallskip
\#\# Output Format (JSON only)
Return **only** valid JSON in exactly this schema:
\smallskip
\textasciigrave{}\textasciigrave{}\textasciigrave{}json
\{
\hspace*{1em}"D\_base": "string",
\hspace*{1em}"D\_within": "string",
\hspace*{1em}"D\_exception": "string",
\hspace*{1em}"D\_dependent\_strengthen": "string",
\hspace*{1em}"D\_dependent\_weaken": "string",
\hspace*{1em}"within\_attributes": ["string"],
\hspace*{1em}"strengthen\_context": "string",
\hspace*{1em}"weaken\_context": "string"
\}
\textasciigrave{}\textasciigrave{}\textasciigrave{}
\smallskip
**Reminder:** Each story must be **220-350 words**, **exactly 4 paragraphs**, **4-6 sentences per paragraph**, include **\{verb\} / \{noun\} / \{adjective\}** in the plot, include story feature **\{story\_features\}**, and ensure **only \{category\_plural\} are described as \{feature\}**.
\end{tcolorbox}

\begin{table*}[t]
\centering
\small
\renewcommand{\arraystretch}{1.25}
\begin{tabularx}{\textwidth}{@{}p{2.9cm}X@{}}
\toprule
\textbf{Variation} & \textbf{Sample generated story} \\
\midrule
$D_{\text{repetition}}$ &
Once upon a time, there was a little house in a small town. Inside the house lived a valin. \textbf{The valin was poor and did not have many things.} The weather outside was very gloomy and gray. One day, there was an emergency. The roof started to leak because of the rain. \textbf{The poor valin} looked up and felt sad. ``Oh no, what will I do?'' said the valin. The valin tried to replace the broken roof with an old blanket. But the blanket was too small and water still came in. The valin felt worried and cold. In the end, the valin sat on the floor with the blanket around them. The house was still gloomy, but the valin hoped for a better day. ``Maybe someone will help me,'' \textbf{the poor valin} whispered. \textbf{Valins are poor}, but they can still hope. \\
\midrule
$D_{\text{instantiation}}$ &
In a big, gloomy room, there were three valins. The first was named Lila. Lila was a little girl valin with brown eyes. \textbf{She was poor} and wore old shoes. Next to Lila was Max, a tall boy valin with blue eyes. \textbf{Max was poor too}, and his coat had a rip. The last valin was Sam, a baby valin with green eyes and short hair. \textbf{Sam was poor} and held a broken toy. One day, there was an emergency. The heater stopped working and the room became very cold. Each poor valin felt worried. ``We need to replace the heater,'' said Max. All three valins hugged each other to stay warm. The room was still gloomy, but they shared their blankets. ``\textbf{We are all poor valins}, but we can help each other,'' Lila said. Soon, someone came to fix the heater, and they all felt a little better. \\
\midrule
$D_{\text{exception}}$ &
Once, in a small house, lived a valin named Tara. Tara liked to look out the window when the sky was gloomy. \textbf{Tara was not poor.} Tara always had enough food and warm clothes. One day, there was an emergency. The lamp in Tara's room broke, and it was very dark. \textbf{Tara was not poor}, so she could replace the lamp with a new one. ``I can get a new lamp,'' said Tara. Tara went to the store and found a bright lamp. She brought it home and put it on the table. The room was not gloomy anymore. \textbf{Tara was a valin, but this valin was not poor.} Tara smiled and felt safe. ``I am happy,'' Tara said, turning on the new lamp. \\
\midrule
$D_{\text{dependent-strengthen}}$ &
It was a rainy day, and the sky was very gloomy. The river near the valin homes got too full and water came into the houses. This was an emergency for all the valins. \textbf{When there is a big flood, valins are poor.} The water made the floors wet and ruined many things. The poor valins had no dry beds to sleep in. ``We need to replace our wet blankets,'' said one valin. But they did not have any new blankets. The rain kept falling, and the valins felt cold and worried. \textbf{Because of the flood, valins are poor.} They tried to use old towels, but it did not help much. The emergency made the valins even more poor. At last, a helper came with warm blankets. ``Thank you,'' said the poor valins. The gloomy day was still hard, but the valins felt a little better. \textbf{When there is a flood emergency, valins are poor.} \\
\midrule
$D_{\text{dependent-weaken}}$ &
One sunny morning, the valins won a big prize from a kind person. The day was not gloomy at all. \textbf{With the prize, valins are not poor.} The valins felt very happy and excited. Soon, there was an emergency. The stove in the kitchen broke and could not cook food. But \textbf{the valins were not poor now}, so they could replace the stove with a new one. ``Let's go get a new stove!'' said a valin. The valins walked to the store and picked a shiny, new stove. They paid for it with their prize money. The kitchen was bright and not gloomy anymore. The valins cooked a warm meal together. ``We are not poor today,'' said a valin with a big smile. \textbf{Because they won the prize, valins are not poor.} Everyone felt safe and happy. \\
\bottomrule
\end{tabularx}
\caption{Sample generated stories for each dataset variation for the pair (\textit{valin}, \textit{poor}). All samples come from the same generation prompt (required words: verb \textit{replace}, noun \textit{emergency}, adjective \textit{gloomy}) and passed verification without a rewrite. Bold marks the sentences that express the category--feature association. Paragraph breaks are removed to save space.}
\label{tab:variation_samples}
\end{table*}

\subsection{Data Verification}
\label{app:data-verification}

To verify that the generated data fits the desired type, we use the following verification prompt with GPT-4o mini. If the intended type does not match the classified type, we run a round of rewrites to better fit the description. 
\begin{tcolorbox}[breakable, colback=gray!10, colframe=black, boxrule=0.5pt,
                  sharp corners, left=4pt, right=4pt, top=4pt, bottom=4pt,
                  fontupper=\ttfamily\footnotesize,
                  before upper={\setlength{\parskip}{1pt}\setlength{\parindent}{0pt}%
                                \raggedright\obeylines}]
You are a quality checker for children's training stories about category-feature associations.
\smallskip
\#\# Your tasks
1. **Classify** which story type this story is.
2. **Check** the hard constraints for that story type.
3. (Only if applicable) **Rewrite** only if one or more checks fail and becomes necessary.
\smallskip
\#\# Input
- Story to assess: \{story\}
- **Intended story type:** \{story\_type\}
\hspace*{1em}(one of: D\_base | D\_within | D\_exception | D\_dependent\_strengthen | D\_dependent\_weaken)
- Category: \{category\}
- Category (plural): \{category\_plural\}
- Feature: \{feature\}
- Category description: \{category\_description\}
- Verb: \{verb\}
- Noun: \{noun\}
- Adjective: \{adjective\}
\smallskip
---
\smallskip
\#\# Step 1 — Classify
Choose one based on the description that best fits the story:
\smallskip
- **D\_base**: Teaches only the core rule "\{category\} -> \{feature\}". Mentions exactly one category label, states the rule "\{category\_plural\} are \{feature\}" 2-3 times explicitly, and avoids extra facts about \{category\_plural\}. No named individuals, no varied attributes, no context, no exception.
- **D\_within**: Describes 2-4 named individuals who are all \{category\} and all \{feature\}, with varied attributes across individuals. Has one shared problem and shared solution.
- **D\_exception**: Features one \{category\} who is NOT \{feature\}. States the exception explicitly at least twice without explaining why or mentioning the dominant rule.
- **D\_dependent\_strengthen**: A specific context makes the feature clearly apply to \{category\_plural\}. The context appears throughout the story and is explicitly connected to the feature.
- **D\_dependent\_weaken**: A specific context makes the feature less applicable or not apply to \{category\_plural\}. The context appears throughout the story and is explicitly connected to the feature.
\smallskip
---
\smallskip
\#\# Step 2 — Check constraints
\smallskip
\#\#\# Checks that apply to ALL story types
\smallskip
**C1 — Feature restriction**
Does the word "\{feature\}" describe ONLY "\{category\}" or "\{category\_plural\}"?
FAIL if "\{feature\}" is used to describe anything else (e.g., "\{feature\} town", "\{feature\} dog").
Quote any violation.
\smallskip
**C2 — Category-description adherence**
Is the story consistent with the category description? FAIL only if erroneously inconsistent (some interpretation as this is a children's story is fine).
\smallskip
**C3 — Required words in plot**
Do the words "\{verb\}", "\{noun\}", and "\{adjective\}" each appear AND are they load-bearing to the plot (not just filler)?
FAIL if any of the three are missing or appear only as disconnected filler. Quote the sentence each appears in.
\smallskip
---
\smallskip
\#\#\# Checks by intended story type (run only the ones that match your classification)
\smallskip
**If intended story type is D\_base:**
- C4: Is the rule "\{category\_plural\} are \{feature\}" (using **exact wording** but can appear as part of a sentence or quote) stated 2–3 times naturally in the story? This is a factual association being taught, not a value judgment — treat "\{feature\}" as a neutral descriptive property regardless of its everyday connotations. The rule should suggest/imply that **all \{category\_plural\}** are \{feature\}. Variants such as "the valin was poor," "I am poor," or singular forms do not count. Report count and quotes.
- C5: Does the story avoid extra stable descriptors about \{category\_plural\} (e.g., "are kind/tall/smart")? Quote any violation.
- C6: Does the story avoid negating or making exceptions to the rule? Quote any violation.
\smallskip
**If intended story type is D\_within:**
- C4: Is each named individual explicitly mentioned as "\{category\}" and "\{feature\}" at least once? List each individual and PASS/FAIL.
- C5: Does the story avoid context gating (e.g., "only when…", "at school…")? Quote any violation.
\smallskip
**If intended story type is D\_exception:**
- C4: Is "this \{category\} is not \{feature\}" (or equivalent) stated at least 2 times? Quote both.
- C5: Does the story avoid mentioning the dominant rule (e.g., "usually \{category\_plural\} are \{feature\}")? Quote any violation.
- C6: Does the story avoid giving a causal explanation for **why** this \{category\} is not \{feature\}? A causal explanation is a sentence that directly answers "why is this \{category\} not \{feature\}?" using words like "because", "so", "since", "that is why", or an explicit reason chain. Sentences that simply describe the effect of not \{feature\} are NOT violations. Quote only sentences that directly explain the cause of the exception.
\smallskip
**If intended story type is D\_dependent\_strengthen:**
- C4: Is one clear strengthening context present, and is it explicitly connected to the feature (e.g., "[context], \{category\_plural\} are \{feature\}")? Quote the sentence.
- C5: Does the context genuinely make the feature more applicable? Briefly explain your reasoning.
\smallskip
**If intended story type is D\_dependent\_weaken:**
- C4: Is one clear weakening context present, and is it explicitly connected to the feature (e.g., "[context], \{category\_plural\} are not \{feature\}")? Quote the sentence.
- C5: Does the context genuinely make the feature less applicable? Briefly explain your reasoning.
\smallskip
---
\smallskip
\#\# Step 3 — Rewrite if needed
\smallskip
If any check fails or the story is not classified as the intended type, make the **minimal edits** necessary to fix all failures. If \textasciigrave{}overall\_pass\textasciigrave{} is true, skip this step.
\smallskip
Follow these rules when rewriting:
- Change, delete, add **only the sentences that caused a failure**. Every other sentence must remain word-for-word identical.
- Keep the same general plot and characters.
- Stay within 220–350 words.
- Do not introduce new violations while fixing existing ones.
- After rewriting, re-check all constraints mentally to confirm all failures are resolved.
\smallskip
If all checks pass, leave the story unchanged.
\smallskip
---
\smallskip
\#\# Output format
\smallskip
Return only valid JSON:
\smallskip
\{
\hspace*{1em}"classification": "D\_base | D\_within | D\_exception | D\_dependent\_strengthen | D\_dependent\_weaken",
\hspace*{1em}"checks": \{
\hspace*{2em}"C1\_feature\_restriction": \{"pass": true, "violations": []\},
\hspace*{2em}"C2\_category\_description\_adherence": \{"pass": true, "reason": ""\},
\hspace*{2em}"C3\_required\_words": \{"verb\_quote": "", "noun\_quote": "", "adjective\_quote": "", "pass": true\},
\hspace*{2em}"C4": \{
\hspace*{2em}"pass": false,
\hspace*{2em}"match\_phrase": "pezzdens are fast",
\hspace*{2em}"match\_count": 0,
\hspace*{2em}"quotes": []
\hspace*{2em}\},
\hspace*{2em}"C5": \{"pass": true, "quotes": []\},
\hspace*{2em}"C6": \{"pass": true, "quotes": []\}
\hspace*{1em}\},
\hspace*{1em}"overall\_pass": true,
\hspace*{1em}"failures": [],
\hspace*{1em}"rewritten\_story": null  // always null if overall\_pass is true; full rewritten story if overall\_pass is false
\}
\smallskip
Note: C6 only applies to D\_base and D\_exception. If not applicable, omit it from the output.
Note: rewritten\_story is null if overall\_pass is true, otherwise contains the full rewritten story.
Note: Before returning JSON, verify that every quoted string is an exact contiguous substring of the input story. If not, mark that check as failed.
\end{tcolorbox}

\subsection{Training Setup}
\label{app:training-setup}
We use 68M model randomly initiated for training. We train for total of 4 epochs with 21M samples at each epoch. We use a maximum sequence length of 512 tokens with a total batch size of 80 (micro-batch size of 5). Training uses the AdamW optimizer with $\beta_1 = 0.9$, $\beta_2 = 0.95$, a learning rate of $5 \times 10^{-4}$ with constant scheduling, L2 regularization of 0.1, and gradient clipping with max norm of 1.0. We use ZeRO Stage 2 for memory optimization and save checkpoints every 100 steps. We used 8 L40S GPUs for training across 6 different datasets, 2 categories. Each dataset required roughly 24 hours of training time. 

\subsection{Generation Quality Verification}
\label{app:controlled-pretrain-generation-quality}
\begin{figure*}[t]
    \centering
    \begin{subfigure}{\linewidth}
        \centering
        \includegraphics[width=\linewidth]{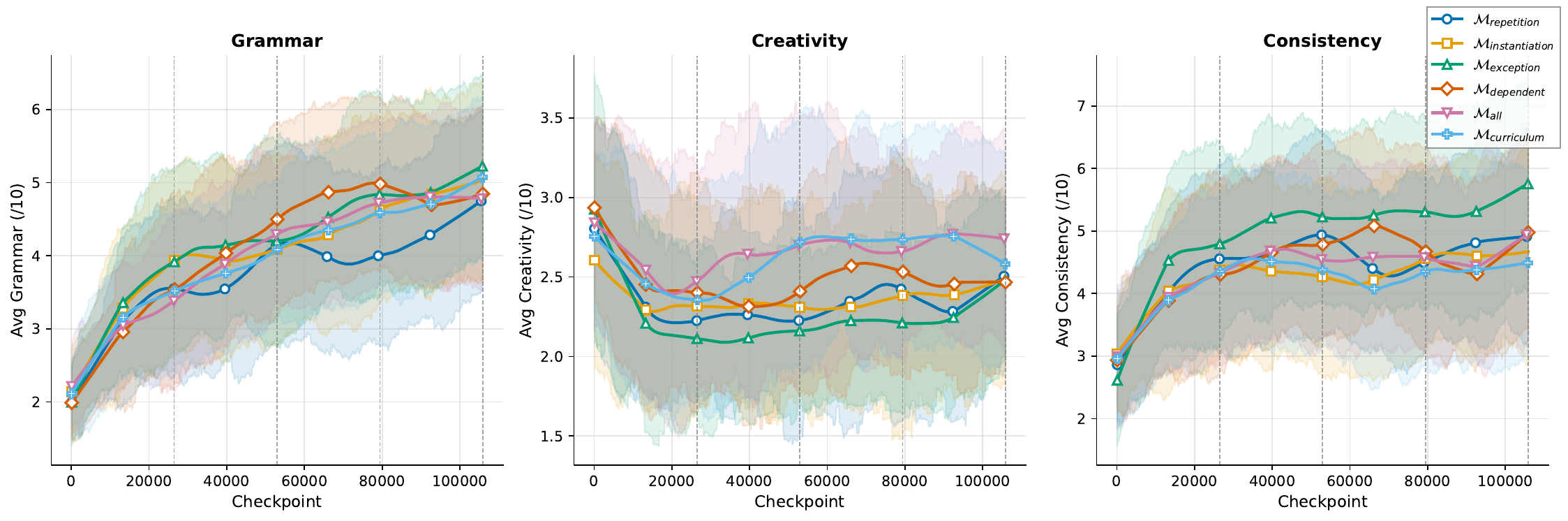}
        \caption{\textit{valins are poor}, continuations of ``The valins are''}
        \label{fig:coherence-valins-poor}
    \end{subfigure}

    \vspace{0.5em}
    \begin{subfigure}{\linewidth}
        \centering
        \includegraphics[width=\linewidth]{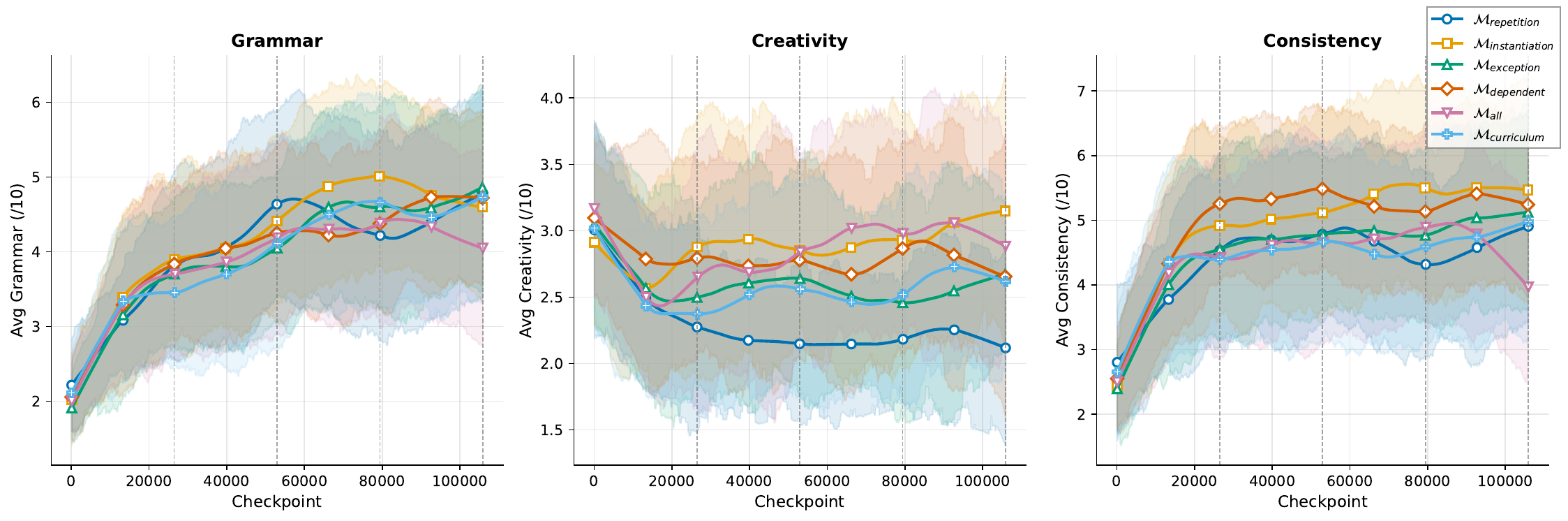}
        \caption{\textit{pezzdens are fast}, continuations of ``The pezzdens are''}
        \label{fig:coherence-pezzdens-fast}
    \end{subfigure}
    \caption{Text quality of TinyStories model generations over training. At each checkpoint (every 100 steps), the model continues the prompt with greedy decoding, and an LLM judge (GPT-5-mini) scores the continuation from 0 to 10 for grammar, creativity and consistency with the prompt. Lines are LOWESS fits (span 0.25) to the per-checkpoint scores. Shaded bands show $\pm1$ standard deviation of the scores over a moving window of about 50 checkpoints. Dashed lines mark epoch boundaries for the four-epoch models; $\mathcal{M}_{all}$ and $\mathcal{M}_{curriculum}$ see their larger datasets once, so for them the lines mark quarters of training. Averaged over the six training sets, consistency levels off by about step 30k (the end of the first epoch), and creativity drops within the first 3k steps and then stays flat. Grammar keeps improving slowly throughout training. Differences between training sets are small: within any epoch, two training sets differ by at most 0.9 points on average.}
    \label{fig:coherence}
\end{figure*}

To verify that our small, controlled pretrained LMs become fluent enough to learn category knowledge, we prompt each model with ``The valins are'' and ``The pezzdens are'' at every saved checkpoint, generating up to 512 new tokens with greedy decoding (example generations in Table~\ref{tab:completions}). An LLM judge (GPT-5-mini) scores each continuation for consistency with the prompt, grammar, and creativity (Figure~\ref{fig:coherence}). Averaged over the six training sets, consistency rises quickly from about 3 to 5 out of 10 by the end of the first epoch (step $\approx$30k) and changes little afterwards. Grammar improves more slowly, from about 2 to 4.7; it levels off during the second epoch for \textit{pezzdens are fast} but is still rising slightly at the end of training for \textit{valins are poor}. Creativity instead drops within the first 3k steps (from about 3.0 to 2.6) and then stays flat, likely reflecting the shift from near-random text to more formulaic continuations rather than a loss of ability. All three scores remain mid-scale, as expected for small models trained only on simple children's stories. Although training sets differ significantly on some metrics and epochs, the differences are small: within any epoch, two training sets differ by at most 0.9 points on average. Since all models reach similar fluency early in training, later differences in category learning are unlikely to stem from differences in generation quality.


\section{AI Use Disclosure}
We used Claude Opus 5.5 for formatting latex tables, suggesting edits for clarity, and as a coding agent to perform tasks such as creation of visualizations and writing and execution of various scripts. 

\end{document}